%% file: main.tex
\documentclass[sigconf]{acmart}
\usepackage{graphicx} 
\usepackage{colortbl}  
\usepackage{xcolor}
\usepackage{siunitx}
\usepackage{enumitem}
\usepackage[normalem]{ulem}
\usepackage{booktabs}
\usepackage{adjustbox}
\usepackage[utf8]{inputenc} 
\usepackage[T1]{fontenc}    
\usepackage{hyperref}       
\usepackage{url}            
\usepackage{amsfonts}       
\usepackage{nicefrac}       
\usepackage{microtype}      
\usepackage{amsmath}
\usepackage{mathtools}
\usepackage{multirow}
\usepackage{makecell}
\usepackage{pifont}
\usepackage[misc]{ifsym}
\usepackage{array}
\usepackage{tabularx}
\usepackage{subcaption} 
\usepackage{wrapfig}
\usepackage{float}
\usepackage{array}
\usepackage{stfloats}
\usepackage{balance}
\usepackage[most]{tcolorbox}

\usepackage{varwidth}
\usepackage{placeins}
\usepackage{amsthm}
\usepackage{algorithm}
\usepackage{algorithmic}

\usepackage[capitalize,noabbrev]{cleveref}

\copyrightyear{2026}
\acmYear{2026}
\setcopyright{cc}
\setcctype{by}
\acmConference[CIKM '26]{Proceedings of the 35th ACM International Conference on Information and Knowledge Management}{November 07--11, 2026}{Rome, Italy}
\acmBooktitle{Proceedings of the 35th ACM International Conference on Information and Knowledge Management (CIKM '26), November 07--11, 2026, Rome, Italy}
\acmDOI{10.1145/3799682.3840596}
\acmISBN{979-8-4007-2539-5/2026/11}

\begin{document}

\title{AlphaCast: A Human Wisdom-LLM Intelligence Co-Reasoning Framework for Time Series Forecasting}


\author{Xiaohan Zhang}
\orcid{0009-0000-4673-4964}
\email{xhzhang1127@mail.ustc.edu.cn}
\affiliation{%
  \institution{State Key Laboratory of Cognitive Intelligence, University of Science and Technology of China}
  \city{Hefei}
  \country{China}
}

\author{Tian Gao}
\orcid{0009-0007-4246-4590}
\email{ustc25gt@mail.ustc.edu.cn}
\affiliation{%
  \institution{State Key Laboratory of Cognitive Intelligence, University of Science and Technology of China}
  \city{Hefei}
  \country{China}
}

\author{Mingyue Cheng}
\orcid{0000-0001-9873-7681}
\authornote{Corresponding author.}
\email{mycheng@ustc.edu.cn}
\affiliation{%
  \institution{State Key Laboratory of Cognitive Intelligence, University of Science and Technology of China}
  \city{Hefei}
  \country{China}
}

\author{Bokai Pan}
\orcid{0009-0001-4496-8166}
\email{bkpan@mail.ustc.edu.cn}
\affiliation{%
  \institution{State Key Laboratory of Cognitive Intelligence, University of Science and Technology of China}
  \city{Hefei}
  \country{China}
}

\author{Ze Guo}
\orcid{0009-0006-9461-5007}
\email{gz1504921411@mail.ustc.edu.cn}
\affiliation{%
  \institution{State Key Laboratory of Cognitive Intelligence, University of Science and Technology of China}
  \city{Hefei}
  \country{China}
}

\author{Yaguo Liu}

\orcid{0009-0006-1668-3696}
\email{ygliu@mail.ustc.edu.cn}
\affiliation{%
  \institution{State Key Laboratory of Cognitive Intelligence, University of Science and Technology of China}
  \city{Hefei}
  \country{China}
}

\author{Xiaoyu Tao}
\orcid{0009-0000-0634-6254}
\email{txytiny@mail.ustc.edu.cn}
\affiliation{%
  \institution{State Key Laboratory of Cognitive Intelligence, University of Science and Technology of China}
  \city{Hefei}
  \country{China}
}

\author{Qi Liu}
\orcid{0000-0001-6956-5550}
\email{qiliuql@ustc.edu.cn}
\affiliation{%
  \institution{State Key Laboratory of Cognitive Intelligence, University of Science and Technology of China}
  \city{Hefei}
  \country{China}
}
\renewcommand{\shortauthors}{Xiaohan Zhang et al.}

\begin{abstract}

\input{0-Abstract.tex}
\end{abstract}


\maketitle

\input{1-Introduction_0523}
\input{2-relatedwork-new}
\input{3-SystemDesign}
\input{4-Experiments}

\input{5-Conclusion}




\clearpage
\appendix
\input{6-Appendix_new}

\clearpage
\section*{GenAI Usage Disclosure}


Generative AI models were used as experimental components of AlphaCast and for limited writing assistance. GPT-5 served as the backbone for the Investigator, Generator, and Reflector, while DeepSeek-V3, DeepSeek-R1, GPT-4o, and Gemini-2.5 Pro were used only in backbone comparison experiments. The authors designed the methodology, prompts, and evaluation protocols and verified all results. GenAI was not used to generate training or evaluation data or ground-truth annotations. Writing assistance was limited to grammar, wording, and readability improvements. The authors take full responsibility for the manuscript and its conclusions.

\bibliographystyle{ACM-Reference-Format}
\balance
\bibliography{main}








\end{document}

%% file: 0-Abstract.tex
Time series forecasting is typically treated as a single-pass numerical prediction task based on historical observations. However, real-world forecasting often requires expert-like reasoning, where temporal patterns, contextual cues, case-based references, and domain knowledge are jointly considered and iteratively revised. Existing statistical, deep learning, foundation, and LLM-based forecasting models mainly rely on pattern fitting, direct prompting, or static context augmentation, leaving interactive and reflective forecasting insufficiently explored. In this work, we propose AlphaCast, a human wisdom–LLM intelligence co-reasoning framework for interactive time series forecasting. AlphaCast does not require real-time human intervention; instead, it operationalizes human wisdom as structured forecasting support, including time-series features, contextual cues, case-based references, and domain knowledge. Built upon an expert-like think–forecast–reflect workflow, AlphaCast comprises three agents: an Investigator that constructs forecasting context from available evidence, a Generator that produces context-grounded forecasts, and a Reflector that evaluates plausibility and provides feedback for refinement. Extensive experiments demonstrate that AlphaCast improves forecasting accuracy and produces interpretable forecast explanations, offering a promising step toward interactive and context-aware time series forecasting\footnote{\url{https://github.com/ustc-time-series/AlphaCast}}.


%% file: 1-Introduction_0523.tex
\section{Introduction}
\begin{figure}
    \centering
    \includegraphics[width=\linewidth]{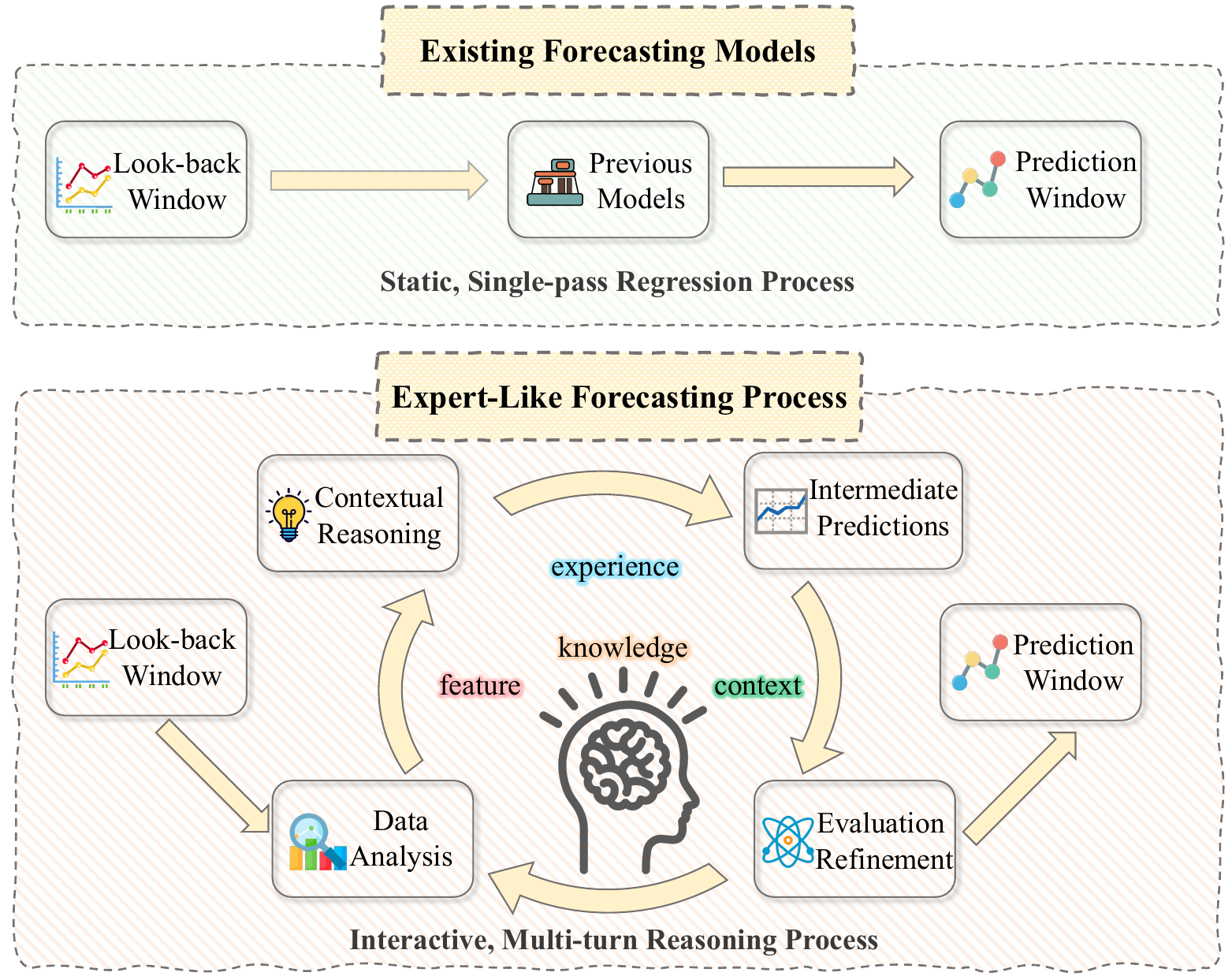}
    \caption{Main idea of AlphaCast: transforming static, single-pass forecasting into expert-like interactive reasoning with analysis, prediction, and refinement.}
    \label{fig:example}
\end{figure}

Time series forecasting supports decision-making in a wide range of real-world applications, including energy scheduling~\cite{zhu2023eforecaster}, climate risk assessment~\cite{dimri2020time}, and industrial operations~\cite{wen2022robust}. In these scenarios, predictions are not merely numerical outputs but operational inputs that guide planning, intervention, and risk control. Modern forecasting systems are often deployed under distributional shifts~\cite{chen2024calibration}, external drivers~\cite{salinas2020deepar, de2023incorporating}, and operational constraints~\cite{wang2025post}, where accurate prediction requires more than fitting historical patterns~\cite{cheng2025comprehensive}. In real-world forecasting, future dynamics are often shaped by temporal characteristics, prediction-window conditions, analogous historical cases, and domain-specific constraints. Human experts therefore rarely rely on a single predictive rule. Instead, they examine time-series patterns, interpret contextual cues, refer to similar historical cases, and consult domain knowledge before revising their judgments through iterative assessment~\cite{arvan2019integrating}. These forecasting behaviors suggest that real-world forecasting is inherently an interactive and context-aware reasoning process rather than a static regression task, a perspective that remains insufficiently reflected in current algorithmic forecasting paradigms~\cite{cheng2026position}.

Despite decades of progress, as shown in the left part of Figure~\ref{fig:example}, time series forecasting is still predominantly formulated as a static, single-pass mapping from historical observations to future values. This paradigm underlies classical statistical models~\cite{anderson2011statistical}, modern deep learning methods~\cite{torres2021deep}, and recent time series foundation models~\cite{liang2024foundation}. Although effective on controlled benchmarks, these approaches provide limited support for explicit context construction, external information use, and forecast revision~\cite{box2015time,li2024deep}. In real-world settings, forecasting often requires contextual grounding, domain knowledge, case-based references, and reflective validation to accommodate changing external influences. The lack of such interaction and reflection may limit adaptability and trustworthiness under complex conditions~\cite{gama2014survey,mellers2024human}. These limitations motivate viewing forecasting as an interactive reasoning process rather than a purely single-pass numerical prediction task.

We argue that time series forecasting should be reformulated as an interactive, multi-turn reasoning process rather than a static, single-pass regression task. In this view, predictions emerge not only from a predictive function, but also from structured context preparation, multi-source information integration, and iterative validation. A forecasting system should be able to identify task-relevant time-series features, ground predictions in contextual cues~\cite{ho2013context}, retrieve case-based references from  historical instances~\cite{ning2025ts,ravuru2024agentic}, incorporate domain knowledge~\cite{cheng2024integrating}, and revise intermediate forecasts through reflective evaluation~\cite{mellers2024human,tao2026memcast}. Such capabilities mirror important behaviors of experts and are essential for operating in complex and dynamic environments. This perspective reframes forecasting as an agentic reasoning problem, where intelligent systems actively prepare, generate, evaluate, and refine predictions rather than passively outputting numerical sequences.

In this work, we propose AlphaCast, a human wisdom--LLM intelligence co-reasoning framework that operationalizes the above forecasting perspective into a concrete agentic forecasting system. AlphaCast does not require real-time human intervention; instead, it represents human wisdom as structured forecasting support, including time-series features, contextual cues, case-based references, and domain knowledge. Built upon an expert-like think--forecast--reflect workflow, AlphaCast organizes forecasting into three stages: autonomous context preparation, context-grounded generative forecasting, and reflective validation. Specifically, the Investigator prepares a structured forecasting context by interacting with the feature set, context repository, case library, and knowledge base. The Generator produces an intermediate forecast through generative reasoning over the prepared context. The Reflector then evaluates forecast plausibility, diagnoses inconsistencies, and provides feedback for refinement. Through this coordinated process, AlphaCast transforms forecasting from a static prediction task into an interactive and context-aware reasoning process.

AlphaCast fundamentally differs from existing LLM-based and foundation-model-based forecasting approaches. Most prior methods treat LLMs or pretrained foundation models as enhanced sequence predictors, mainly improving how temporal inputs are represented, prompted, or generated within a single forecasting pass. AlphaCast instead focuses on how the forecasting process itself is organized: the LLM is placed within a structured workflow where forecasting context is prepared before generation and intermediate predictions are assessed before final output. This process-level design enables AlphaCast to move beyond direct sequence prediction toward interactive and context-aware forecasting.
The contributions of this work are as follows:

\begin{itemize}
    \item We reformulate time series forecasting as an interactive reasoning process and propose AlphaCast, a human wisdom--LLM intelligence co-reasoning framework that organizes forecasting into a think--forecast--reflect 
    workflow.

    \item We design a lightweight forecasting toolkit within AlphaCast, integrating time-series features, contextual cues, case-based references, and domain knowledge to construct comprehensive forecasting contexts for LLM-based reasoning.
    \item We conduct extensive experiments on both short-term and long-term real-world forecasting benchmarks, demonstrating consistent improvements over leading statistical, deep learning, foundation, and LLM-based forecasting baselines.
\end{itemize}

%% file: 2-relatedwork-new.tex
\section{Related Work}
\subsection{Conventional Time Series Forecasting}

Time series forecasting has evolved from classical statistical modeling to deep learning architectures and, more recently, foundation models~\cite{qiu2024tfb,cheng2025comprehensive}. Classical statistical methods, including Seasonal Naive, ARIMA~\cite{ARIMA}, CES~\cite{svetunkov2022complex}, and Prophet~\cite{taylor2018forecasting}, provide simple and interpretable solutions by modeling temporal regularities with predefined structures or assumptions. Machine learning methods further extend this line by incorporating lagged observations, covariates, and engineered temporal features, while still relying on task-specific feature design.
Deep learning methods substantially reduce the reliance on manual feature engineering and broaden the architectural design space of forecasting models~\cite{eldele2021tstcc}. Existing studies have explored diverse architectures~\cite{wang2024deep,li2025hyperimts,huang2025timebase}, including linear models, convolution-based models~\cite{eldele2024tslanet}, Transformer variants~\cite{chen2025closer,zhou2022fedformer}, and state-space models~\cite{ahamed2024timemachine}. Representative models such as N-HiTS~\cite{challu2023nhits}, ETSformer~\cite{woo2022etsformer}, iTransformer~\cite{liu2023itransformer}, and ConvTimeNet~\cite{cheng2025convtimenet} show that competitive forecasting performance can emerge from different architectural biases rather than a single dominant backbone. More recently, foundation models such as Chronos~\cite{ansari2024chronos}, TimesFM~\cite{das2023timesfm}, and Sundial~\cite{liu2025sundial} have demonstrated promising zero-shot and cross-domain forecasting ability through large-scale pretraining. Despite their strong performance, most of these methods still perform forecasting in a single-pass manner, which is different from the iterative reasoning process commonly used in real-world forecasting.

\subsection{LLM-Based Time Series Forecasting}

Recent LLM-based forecasting studies leverage the instruction-following, generalization, and reasoning capabilities of language models for time series forecasting. A major line of work focuses on bridging temporal signals and pretrained LLMs: Time-LLM~\cite{jin2024time} reprograms frozen LLMs with temporal representations, while LSTPrompt~\cite{liu2024lstprompt} explores prompt-based forecasting adaptation. These methods demonstrate the value of pretrained language-model priors, but forecasting remains largely organized as direct generation or prediction.
More recent work explores reasoning-enhanced forecasting. TimeReasoner~\cite{wang2025can} introduces explicit temporal reasoning, Time-R1~\cite{wang2026time} strengthens multi-step reasoning through reinforcement fine-tuning, and TSci~\cite{zhao2025timeseriesscientist} organizes LLMs into a structured time series analysis process. Although these studies move beyond direct sequence prediction, reasoning is often completed within a single forecasting pass, without an explicit mechanism to validate intermediate forecasts and use feedback for refinement.
This direction is related to agentic LLM research, where Toolformer~\cite{schick2023toolformer}, ReAct~\cite{yao2023react}, and DEPS~\cite{wang2024describeexplainplanselect} demonstrate tool use, external interaction, and planning in process-oriented systems. However, applying such mechanisms to forecasting requires designs that ensure both contextual grounding and numerical reliability. This motivates forecasting frameworks that treat prediction as a structured, iterative process rather than a single-step generation task.

%% file: 3-SystemDesign.tex
\begin{figure*}
    \centering
    \includegraphics[width=\linewidth]{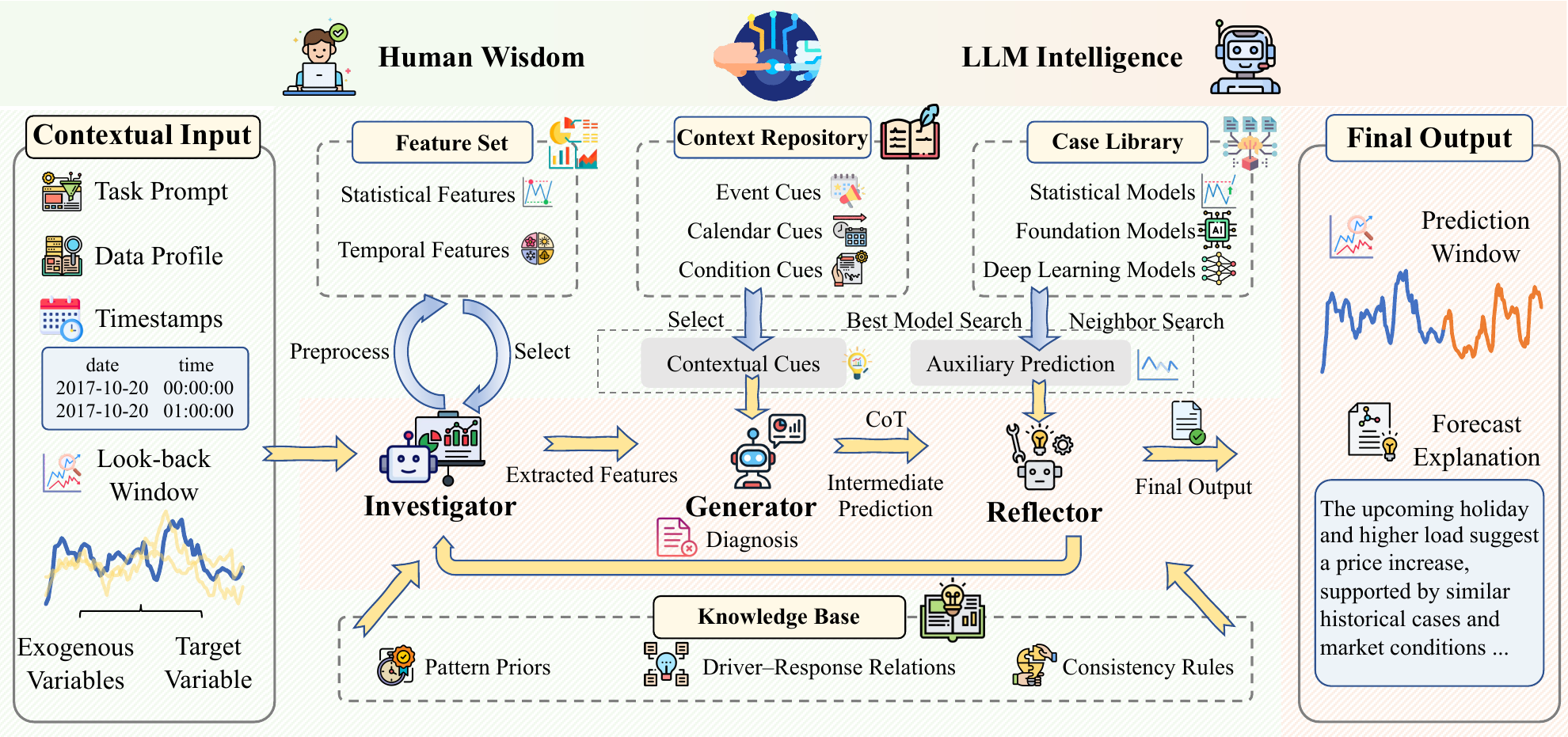}
    \caption{Overall architecture of AlphaCast. The three-stage process (Investigator, Generator, Reflector) bridges human wisdom and LLM intelligence, simulating expert cognition to transform forecasting into an interactive, multi-turn reasoning process.}
    \label{fig:framework}
\end{figure*}
\section{The Proposed AlphaCast}
\label{sec:method}

In this section, we introduce AlphaCast, which reformulates forecasting as a multi-stage process that simulates expert-like reasoning to prepare, generate, and evaluate predictions.

\subsection{Problem Definition}
In forecasting with exogenous variables, the input consists of a historical target series and its associated exogenous information. Let \(X_{\text{en}} \in \mathbb{R}^{H}\) denote the target series observed over a look-back window of length \(H\). Let \(X_{\text{ex}} \in \mathbb{R}^{D\times(H+L)}\) denote the associated exogenous variables, which comprise multiple exogenous series \(\{X_{\text{ex}}^{(1)}, X_{\text{ex}}^{(2)}, \ldots, X_{\text{ex}}^{(D)}\}\). Specifically, \(X_{\text{ex}}^{(d)} \in \mathbb{R}^{H+L}\) denotes the \(d\)-th exogenous series, including its historical values in the look-back window and its known values over the prediction window. 
The goal is to estimate the future target series 
\(Y \in \mathbb{R}^{L}\) over the next \(L\) time steps, and the framework output is denoted as \(\hat{Y} \in \mathbb{R}^{L}\).



\subsection{Framework Overview}
As shown in Figure~\ref{fig:framework}, AlphaCast is a multi-stage agentic reasoning framework for time series forecasting with training-free LLMs. It constructs a lightweight forecasting environment composed of a feature set, a contextual repository, a case library, and a knowledge base, which provides forecasting-relevant information beyond the raw numerical time series. The forecasting process is then organized into three stages: the Investigator prepares a structured forecasting context, the Generator produces an intermediate prediction through reasoning-based generation, and the Reflector validates and refines the prediction before final output.

\subsection{Toolkit Design}
To support autonomous interactive forecasting, we construct a lightweight forecasting environment that provides forecasting-relevant auxiliary information beyond the raw historical time series.


\subsubsection{Feature Set.}
AlphaCast extracts time-series features from both the target time series and available exogenous variables to summarize temporal dynamics. Given a group of extraction functions $\mathcal{F} = \{f_1, \dots, f_n\}$, we define the collection of input time series as
$\mathcal{X} = \{X_{\text{en}}\} \cup \{X_{\text{ex}}^{(d)}\}_{d=1}^{D}$.
The feature set $F$ is constructed by applying each extraction function to each time series in $\mathcal{X}$:
\begin{equation}
F = \{ f(X) \mid f \in \mathcal{F},\; X \in \mathcal{X} \},
\end{equation}
resulting in twenty feature types that characterize the input series from complementary perspectives. Specifically, statistical features summarize distributional properties, including value range, skewness, and kurtosis, whereas temporal features capture dynamic patterns through autocorrelation, differenced dependence, spectral entropy, and seasonality. Together, these features provide compact numerical descriptions that support subsequent context preparation and reasoning-based forecasting. Details of all features are provided in Appendix~\ref{app:feature_set}.

\subsubsection{Contextual Repository}
The contextual repository stores cues that describe the background of the current prediction window and provide instance-specific information for forecasting. These cues include event cues, such as scheduled or abnormal events around the prediction window (e.g., market events, weather alerts, or policy changes); calendar cues, including holidays, weekends, working days, and periodic positions; and condition cues describing the prevailing background state, such as a normal positive-price regime in an electricity market. Forecast-window covariates and cues are used only when released no later than the forecast origin; no future observation, target value, or post-hoc condition summary is used, and the same admissible covariates are supplied to all methods.

\subsubsection{Case Library}
\label{sec:case_library}

The case library stores historical forecasting instances and provides case-based references for the current prediction task. Each training sample \(\langle X_i, Y_i \rangle\) is regarded as a historical case, where \(X_i\) denotes the look-back input and \(Y_i\) denotes the corresponding prediction window. Each \(M_i\) is selected by rolling-origin out-of-sample loss, with candidate models fit only on observations preceding the case origin. The library contains only earlier cases; flattened look-backs are z-scored and clustered by Euclidean K-means (fixed seed; \(K\) selected on validation). Let \(C_m\) denote the index set of samples assigned to the \(m\)-th cluster, and let \(c_m\) denote its cluster center.
During inference, given the current look-back input \(X_{\mathrm{en}}\), AlphaCast first identifies the nearest cluster:
\begin{equation}
m^* = \arg\min_m \operatorname{Dist}\left(X_{\mathrm{en}}, c_m\right),
\end{equation}
where \(\operatorname{Dist}(\cdot,\cdot)\) measures the similarity between look-back sequences. Based on the selected cluster, AlphaCast obtains a cluster-based prediction by aggregating the outputs of the candidate models associated with its member cases:
\begin{equation}
Y_{\mathrm{case}} =
\frac{1}{|C_{m^*}|}
\sum_{i \in C_{m^*}} M_i(X_{\mathrm{en}}),
\end{equation}
which is a frequency-weighted ensemble: a candidate model contributes once for every historical case in the selected cluster for which it achieved the lowest out-of-sample loss.

To provide a more concrete historical analogy, AlphaCast also retrieves the nearest historical case from the entire training set:
\begin{equation}
i^* = \arg\min_{1 \leq i \leq N}
\operatorname{Dist}\left(X_{\mathrm{en}}, X_i\right).
\end{equation}
The retrieved case is denoted as \((X_{\mathrm{nb}}, Y_{\mathrm{nb}})\), where \(X_{\mathrm{nb}} = X_{i^*}\) is the neighbor look-back sequence and \(Y_{\mathrm{nb}} = Y_{i^*}\) is its corresponding future series. Therefore, the case library provides two complementary references for subsequent reasoning: \(Y_{\mathrm{case}}\) summarizes cluster-level forecasting tendencies, while \((X_{\mathrm{nb}}, Y_{\mathrm{nb}})\) offers an instance-level input-output analogy for prediction adjustment and refinement.
\subsubsection{Knowledge Base}

AlphaCast maintains a knowledge base that provides reusable prior knowledge to support reasoning and decision-making. Let $K$ denote the domain knowledge selected from the knowledge base. It contains pattern priors, such as periodicity and fluctuation, which provide general expectations about plausible target-series evolution; driver--response relations, which characterize how external drivers or contextual conditions affect the direction, magnitude, and timing of prediction adjustments; and consistency rules, including valid value ranges, temporal causality, and logically consistent responses.

\subsection{Contextual Input Construction}

Accurate forecasting requires a clear task specification, explicit variable descriptions, and temporally aligned observations. AlphaCast therefore constructs a standardized contextual input \(I_{\text{in}}\) that combines textual and numerical inputs:
\begin{equation}
I_{\text{in}} = (P_{\text{task}}, D_{\text{profile}}, T, X_{\text{en}}, X_{\text{ex}}),
\end{equation}
where the components are organized by input form to support structured context construction.

The textual inputs provide task-level descriptions and semantic information that help the LLM interpret the forecasting task before reasoning over numerical values. Specifically, the task prompt \(P_{\text{task}}\) specifies the look-back length, prediction horizon, and required output format, aligning generation with the forecasting objective. The data profile \(D_{\text{profile}}\) describes the target and exogenous variables, including their domain meanings, units, and relationships to the target. Timestamps \(T\) specify the temporal positions of the look-back and prediction windows, enabling the model to recognize both absolute time information, such as dates and hours, and relative temporal patterns, such as periodicity and seasonality~\cite{wang2024rethinking,zeng2024much}.

The numerical inputs consist of the historical target series \(X_{\text{en}}\) and the exogenous variables \(X_{\text{ex}}\). The target series represents the endogenous sequence in the look-back window, providing the primary temporal evidence for identifying recent trends, fluctuations, and local variations. The exogenous variables include available external signals over the look-back and prediction windows, supplying additional evidence about external influences without accessing future target values. Together, these inputs provide the basis for subsequent context construction, reasoning-based generation, and reflective refinement.

\subsection{Forecasting Context Preparation}

Given the standardized contextual input \(I_{\text{in}}\), the Investigator prepares a structured forecasting context for subsequent reasoning. It first interprets the task prompt \(P_{\text{task}}\), data profile \(D_{\text{profile}}\), timestamps \(T\), and available exogenous variables to clarify the forecasting target, prediction horizon, variable semantics, and usable information within the prediction window. Meanwhile, the Investigator interacts with the forecasting toolkit to understand the auxiliary information available for the current task, including domain knowledge from the knowledge base, supplementary context from the contextual repository, and case-based references from the case library.

Based on this task and context understanding, the Investigator autonomously selects informative time series features from the feature set, denoted as \(F_{\text{sel}}\). It also collects domain knowledge \(K\), supplementary context \(S\), and case-based references, including the cluster prediction \(Y_{\text{case}}\), the nearest historical neighbor \(X_{\text{nb}}\), and its corresponding future series \(Y_{\text{nb}}\). These components serve distinct but complementary roles in context preparation. The selected features describe the statistical and temporal properties of the current input series, domain knowledge provides reusable forecasting priors, contextual cues characterize the prediction scenario, and historical cases offer reference trajectories for comparison and adjustment.
The Investigator integrates these elements into a comprehensive forecasting context:
\begin{equation}
I_{\text{con}} = ( I_{\text{in}}, F_{\text{sel}}, K, S, Y_{\text{case}}, X_{\text{nb}}, Y_{\text{nb}} ).
\end{equation}
This context provides the information basis for subsequent generative reasoning and reflective refinement.



\subsection{Generative Reasoning for Forecasting}

Given the consolidated forecasting context \(I_{\text{con}}\), the Generator aims to transform the prepared forecasting context into an intermediate prediction \(Y_{\text{int}}\). It uses case-based references as the initial prediction basis and adapts them according to the standardized input and auxiliary information organized in \(I_{\text{con}}\). This design allows the generated prediction to inherit useful historical experience while remaining responsive to the current prediction window.

The case-based references provide two complementary forms of historical guidance. The cluster-based prediction \(Y_{\text{case}}\) serves as a coarse forecasting basis. As defined in the case library, \(Y_{\text{case}}\) is obtained by aggregating the predictions of candidate models associated with the matched cluster, and therefore reflects the common predictive tendency of similar look-back patterns. However, this reference may smooth out instance-specific variations and may not fully match the current input. To provide a more concrete historical analogy, the Generator further uses the nearest historical case \((X_{\text{nb}}, Y_{\text{nb}})\). By comparing the nearest historical input \(X_{\text{nb}}\) with the current input \(X_{\text{en}}\), the Generator uses the corresponding future series \(Y_{\text{nb}}\) as an instance-level reference for adapting the cluster prediction~\cite{liu2025improving}. This instance-level reference helps complement the cluster-level prediction with more specific historical evidence.

Beyond case-based references, the Generator uses the remaining information in \(I_{\text{con}}\) to adapt the historical basis to the current task. The input \(I_{\text{in}}\) grounds generation in the task requirements and available observations. The selected features \(F_{\text{sel}}\) help diagnose current statistical and temporal patterns, while domain knowledge \(K\) and contextual cues \(S\) guide plausible adjustments using domain constraints and rich prediction-window information. By integrating this information, the Generator transforms case-based references into a context-aware intermediate prediction \(Y_{\text{int}}\), which is then passed to the Reflector for validation and refinement.
\input{table/main_result_2}

\subsection{Iterative Validation and Refinement}

In real-world forecasting, predictions are often reviewed and refined before being finalized. AlphaCast therefore includes a reflection stage to validate the intermediate prediction and support iterative refinement. This stage enhances the reliability of the framework through two complementary checks. First, prediction evaluation assesses whether the generated forecast satisfies the task requirements and remains consistent with the available contextual information and domain knowledge. Second, reasoning evaluation examines whether the generated explanation uses the forecasting context coherently and remains consistent with the predicted series. Together, these checks help ensure that both the prediction and its explanation align with the task requirements.
If issues are identified, AlphaCast uses the feedback to revise \(Y_{\text{int}}\) and obtains a refined prediction. The refined result is then evaluated using the same criteria. If it is accepted, AlphaCast returns it as the final forecast \(\hat{Y}\) over the target horizon; otherwise, the feedback is returned to the Investigator for another update. This process repeats until an acceptable forecast is obtained or the predefined refinement budget is reached. In addition to the numerical prediction, AlphaCast can optionally provide a concise summary that records the key contextual cues, case-based references, domain knowledge, and major adjustment decisions used during generation and refinement. This summary improves interpretability by explaining the main factors behind the forecast without exposing detailed reasoning traces.

%% file: table/main_result_2.tex
\begin{table*}[t!]
\centering
\small
\caption{Overall forecasting performance under short-term and long-term forecasting on various benchmark datasets. Lower values indicate better performance. The best results are highlighted in \textbf{bold}, and the second-best are \underline{underlined}.}
\label{tab:main_results}
\setlength{\tabcolsep}{3.2pt}
\renewcommand{\arraystretch}{1.18}
\resizebox{\textwidth}{!}{
\begin{tabular}{c|cc cc cc cc cc|cc cc cc cc cc}
\toprule
\multicolumn{1}{c|}{Settings} 
& \multicolumn{10}{c|}{Short-term Forecasting} 
& \multicolumn{10}{c}{Long-term Forecasting} \\
\cmidrule(r){1-1} \cmidrule(lr){2-11} \cmidrule(l){12-21}
\multicolumn{1}{c|}{Dataset} 
& \multicolumn{2}{c}{NP} 
& \multicolumn{2}{c}{PJM} 
& \multicolumn{2}{c}{BE} 
& \multicolumn{2}{c}{FR} 
& \multicolumn{2}{c|}{DE} 
& \multicolumn{2}{c}{ETTh} 
& \multicolumn{2}{c}{ETTm} 
& \multicolumn{2}{c}{WP} 
& \multicolumn{2}{c}{SP} 
& \multicolumn{2}{c}{MOPEX} \\
\multicolumn{1}{c|}{Metric} 
& MSE & MAE 
& MSE & MAE 
& MSE & MAE 
& MSE & MAE 
& MSE & MAE 
& MSE & MAE 
& MSE & MAE 
& MSE & MAE 
& MSE & MAE 
& MSE & MAE \\
\midrule

SNaive 
& 44.89 & 4.10 
& 41.39 & 4.65 
& 857.12 & 13.70 
& 915.80 & 11.15 
& 415.72 & 13.36 
& 10.75 & 2.47 
& 2.75 & 1.19 
& 1948.06 & \underline{27.88} 
& 79.61 & 4.56 
& 7.18 & 1.62 \\

ARIMA 
& 55.51 & 4.45 
& 33.02 & 4.00 
& 869.44 & 17.04 
& 964.00 & 13.22 
& 372.91 & 11.82 
& 10.88 & 2.43 
& 2.71 & 1.17 
& 2166.64 & 28.77 
& 80.48 & 4.60 
& 9.19 & 1.85 \\

CES 
& 40.78 & 3.99 
& 38.33 & 4.47 
& 808.61 & 14.41 
& 1084.71 & 12.14 
& 315.58 & 11.45 
& 24.22 & 3.65 
& 4.12 & 1.48 
& 9364.98 & 42.04 
& 78.71 & 4.55 
& 6.40 & 1.47 \\

Prophet 
& 57.79 & 5.00 
& 52.26 & 5.53 
& 992.90 & 16.94 
& 1035.70 & 14.42 
& 320.63 & 13.25 
& 46.70 & 4.59 
& 21.18 & 3.11 
& 8071.24 & 56.06 
& 64.81 & 6.44 
& 12.39 & 2.50 \\

Autoformer 
& 46.10 & 4.86 
& 77.57 & 6.43 
& 890.84 & 16.62 
& 934.36 & 13.36 
& 331.31 & 13.33 
& 11.60 & 2.64 
& 4.22 & 1.57 
& 2567.61 & 37.00 
& 39.13 & 4.34 
& 6.17 & 1.96 \\

DLinear 
& 32.22 & 3.84 
& 42.15 & 4.64 
& 658.53 & 12.83 
& 811.45 & 10.45 
& 239.93 & 12.83 
& 8.51 & 2.24 
& 2.63 & 1.14 
& 1932.20 & 29.32 
& 19.06 & 2.73 
& 5.26 & 1.33 \\

TimesNet 
& 32.12 & 3.81 
& 34.89 & 4.34 
& 636.66 & 11.30 
& 929.10 & 11.10 
& 209.37 & 10.20 
& \underline{7.94} & \underline{2.08} 
& \underline{2.44} & \underline{1.07} 
& \underline{1930.88} & 29.40 
& 18.43 & 2.78 
& 4.96 & 1.35 \\

PatchTST 
& \underline{24.63} & 3.27 
& 31.87 & 4.17 
& 627.15 & 11.44 
& \underline{797.26} & 9.13 
& \underline{208.89} & 10.04 
& 8.40 & 2.17 
& 2.62 & 1.11 
& 2269.64 & 32.46 
& 19.43 & 2.78 
& 5.36 & 1.44 \\

iTransformer 
& 27.09 & 3.43 
& 35.13 & 4.37 
& \underline{606.53} & 11.24 
& 940.23 & 11.09 
& 229.96 & 10.23 
& 8.31 & 2.14 
& 3.14 & 1.28 
& 2063.99 & 29.89 
& \underline{17.96} & \underline{2.67} 
& 4.88 & 1.34 \\

TimeXer 
& 27.31 & 3.44 
& 25.88 & 3.65 
& 702.86 & 10.70 
& 834.69 & 9.38 
& 252.00 & 10.26 
& 8.77 & 2.20 
& 2.56 & 1.11 
& 2169.82 & 31.91 
& 20.11 & 2.92 
& \underline{4.83} & \textbf{1.28} \\

Chronos 
& \textbf{22.18} & \underline{3.18} 
& \underline{25.70} & \underline{3.63} 
& 625.63 & \underline{9.62} 
& 803.08 & \underline{7.54} 
& 223.15 & \underline{10.00} 
& 9.40 & 2.25 
& 2.75 & 1.18 
& 2268.52 & 29.62 
& 79.25 & 4.57 
& 5.28 & \underline{1.31} \\

Sundial 
& 28.67 & 3.36 
& 30.70 & 3.99 
& 651.24 & 11.00 
& 942.08 & 8.54 
& 264.62 & 10.71 
& 9.44 & 2.31 
& 3.21 & 1.26 
& 2167.94 & 31.22 
& 100.30 & 6.69 
& 5.15 & 1.42 \\

TimeLLM 
& 84.41 & 6.30 
& 157.17 & 10.98 
& 1229.72 & 20.08 
& 1637.21 & 20.58 
& 503.40 & 16.00 
& 10.99 & 2.60 
& 3.77 & 1.42 
& 3177.06 & 39.23 
& 64.82 & 6.18 
& 6.62 & 1.71 \\

LSTPrompt 
& 52.87 & 4.89 
& 123.12 & 7.89 
& 1031.00 & 17.43 
& 1209.12 & 16.41 
& 389.89 & 13.43 
& 14.24 & 2.80 
& 8.22 & 1.88 
& 3100.17 & 34.20 
& 141.12 & 7.00 
& 8.91 & 1.74 \\

TimeReasoner 
& 68.12 & 5.23 
& 54.71 & 4.90 
& 772.83 & 11.99 
& 852.12 & 12.10 
& 331.83 & 12.92 
& 9.87 & 2.28 
& 3.88 & 1.48 
& 2287.11 & 31.98 
& 26.19 & 2.77 
& 7.18 & 1.59 \\

Time-R1 
& 45.84 & 4.76 
& 69.41 & 6.10 
& 869.19 & 13.88 
& 947.89 & 14.76 
& 297.74 & 12.44 
& 8.73 & 2.34 
& 3.72 & 1.39 
& 2578.39 & 34.26 
& 37.80 & 3.52 
& 6.53 & 1.51 \\

TSci
& 138.29 & 8.05 
& 123.01 & 8.45 
& 1106.68 & 18.94 
& 11989.09 & 17.79 
& 345.40 & 13.39 
& 13.99 & 2.66 
& 6.13 & 1.65 
& 6661.12 & 44.12 
& 75.12 & 5.61 
& 14.18 & 2.04 \\

\midrule
\rowcolor[gray]{0.95} 
\textbf{AlphaCast} 
& 27.11 & \textbf{3.15} 
& \textbf{25.15} & \textbf{3.60} 
& \textbf{536.45} & \textbf{9.61} 
& \textbf{721.02} & \textbf{7.13} 
& \textbf{193.83} & \textbf{9.93} 
& \textbf{7.64} & \textbf{2.02} 
& \textbf{2.41} & \textbf{1.06} 
& \textbf{1548.83} & \textbf{24.54} 
& \textbf{13.29} & \textbf{1.84} 
& \textbf{4.77} & 1.35 \\

\bottomrule
\end{tabular}
}
\end{table*}

%% file: 4-Experiments.tex
\section{Experiments}

In this section, we conduct experiments on various forecasting benchmarks comparing AlphaCast with mainstream approaches to demonstrate its effectiveness.
\subsection{Experimental Settings}
\subsubsection{Datasets.} 
We evaluate our framework on a diverse set of benchmarks covering both short-term and long-term horizons with rich contextual dependencies.
For short-term forecasting, we use five regional datasets from the electricity price forecasting (EPF) benchmark~\cite{wang2024timexer}: NP, PJM, BE, FR, and DE. These datasets contain electricity prices and exogenous variables such as generation and load forecasts.
In the long-term setting, the ETTh and ETTm datasets~\cite{zhou2021informer} are utilized to monitor transformer temperature and load variations.
To capture renewable energy dynamics, we adopt the Windy Power (WP) and Sunny Power (SP) datasets~\cite{xfyun_renewable_power_challenge_2025}, which align power generation records with meteorological conditions.
Additionally, the MOPEX dataset~\cite{schaake2006us} is included for hydrological forecasting, featuring streamflow series supported by climatic factors.
Detailed dataset descriptions are provided in Table~\ref{tab:dataset}.

\subsubsection{Baselines.}
We compare our method with seventeen representative baselines from four categories.
For statistical models, we include Seasonal Naive (SNaive)~\cite{hyndman2018forecasting}, ARIMA~\cite{ARIMA}, CES~\cite{svetunkov2022complex}, and Prophet~\cite{taylor2018forecasting}.
For deep learning models, we evaluate Autoformer~\cite{wu2021autoformer}, DLinear~\cite{zeng2023transformers}, TimesNet~\cite{wu2022timesnet}, PatchTST~\cite{nie2022time},  iTransformer~\cite{liu2023itransformer}, and TimeXer~\cite{wang2024timexer}.
For LLM-based forecasting models, we include Time-LLM~\cite{jin2024time}, LSTPrompt~\cite{liu2024lstprompt}, TimeReasoner~\cite{wang2025can}, Time-R1~\cite{wang2026time}, and TSci~\cite{zhao2025timeseriesscientist}.
For foundation models, we also compare with Chronos~\cite{ansari2024chronos} and Sundial~\cite{liu2025sundial}.
These baselines enable a comprehensive and rigorous comparison across classical statistical methods, deep learning models, foundation models, and emerging LLM-based forecasting approaches.

\subsubsection{Implementation Details.} 
We use \texttt{gpt-5-2025-08-07} via the Responses API (temperature 0.6, top-p 0.7, 16,384 maximum tokens, medium reasoning effort). Each instance executes Investigator--Generator--Reflector once, with at most two refinement rounds. Agents return schema-validated JSON; malformed, non-finite, or wrong-horizon outputs are retried once, then replaced by \(Y_{\mathrm{case}}\). We report the mean and standard deviation over three fixed-seed runs.
Deep learning baselines use official PyTorch implementations on one NVIDIA GeForce RTX 4090D GPU.
For short-term forecasting, we adopt the standard protocol of N-BEATSx~\cite{olivares2023neural}, where the input length is 168 and the horizon is 24. 
For long-term forecasting,
we fix the look-back window and prediction length at 96. We report MSE and MAE on raw, unnormalized series for all methods.
\input{table/datasets}

\input{table/ablation}
\subsection{Main Results}

Table~\ref{tab:main_results} compares AlphaCast against seventeen baselines across short-term and long-term benchmarks. AlphaCast achieves the best performance on 18 out of 20 dataset--metric pairs, demonstrating its effectiveness across different forecasting horizons and benchmark scenarios.
In short-term scenarios, AlphaCast shows outstanding stability in the face of high volatility, effectively managing unpredictable fluctuations. Unlike statistical methods such as ARIMA, which struggle with rapid non-linear fluctuations and inherent stochastic volatility in price dynamics, or deep learning models such as PatchTST, which may over-smooth dynamics, AlphaCast effectively captures sharp peaks by leveraging historical references to avoid phase shifts. This gives it a significant advantage on datasets like BE and FR.
In long-term settings, AlphaCast exhibits superior forecasting capability, especially on datasets like WP and SP, which exhibit complex environmental dependencies. While foundation models such as Chronos perform competitively due to their zero-shot generalization, AlphaCast maintains strong performance through reasoning and refinement guided by domain-informed temporal and consistency constraints. For instance, on the WP dataset, AlphaCast substantially outperforms baselines like DLinear.
This superiority across different scales supports the effectiveness of the interactive, multi-turn reasoning process in improving empirical forecasting accuracy and reliability.
\begin{figure}
    \centering
    \includegraphics[width=\linewidth]{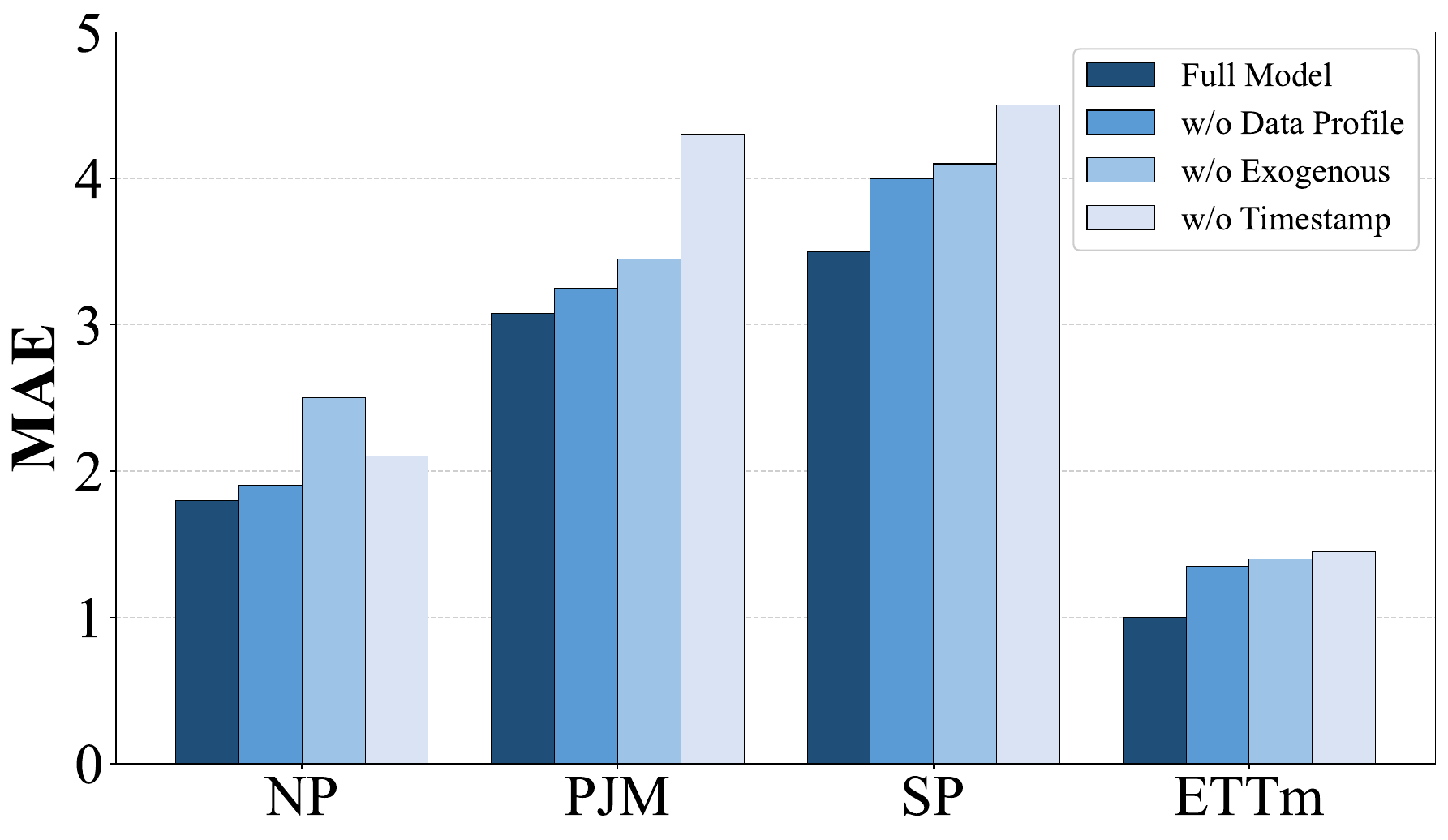}

\caption{Component-wise ablation of contextual input and its impact on model performance.}
    \label{fig:ablation_context}
\end{figure}


\subsection{Ablation Studies}
This section validates the AlphaCast framework through a multi-perspective ablation study, 
covering both architectural stages, contextual input, and auxiliary toolkit modules.

\subsubsection{Ablation Study on Framework Components.}
To validate the workflow, we conduct an ablation study by removing the toolkit, reasoning, and reflection. As shown in Table~\ref{tab:ablation_full}, the full model consistently achieves the best performance, demonstrating the complementary effects of these components. Specifically, removing the toolkit yields the highest errors in scenarios like BE and ETTm, confirming that external context is foundational for accurate forecasting. Furthermore, in the variant without reasoning, the Generator directly uses the cluster-based prediction as its output, without performing context-grounded generative adjustment. Comparing this variant with the full model shows that reasoning-based forecasting substantially improves accuracy, as evidenced by the notable error reduction on the FR dataset. Most importantly, results on the SP dataset highlight the importance of the reflection module, where the variant excluding reflection performs worse than even the non-reasoning variant. This anomaly suggests that without reflection, generative reasoning may introduce unreliable adjustments or overfit to partial contextual evidence. This issue is mitigated by the full model, where reflective validation serves as a safeguard for reliability and domain consistency.

\begin{figure}
    \centering
    \includegraphics[width=\linewidth]{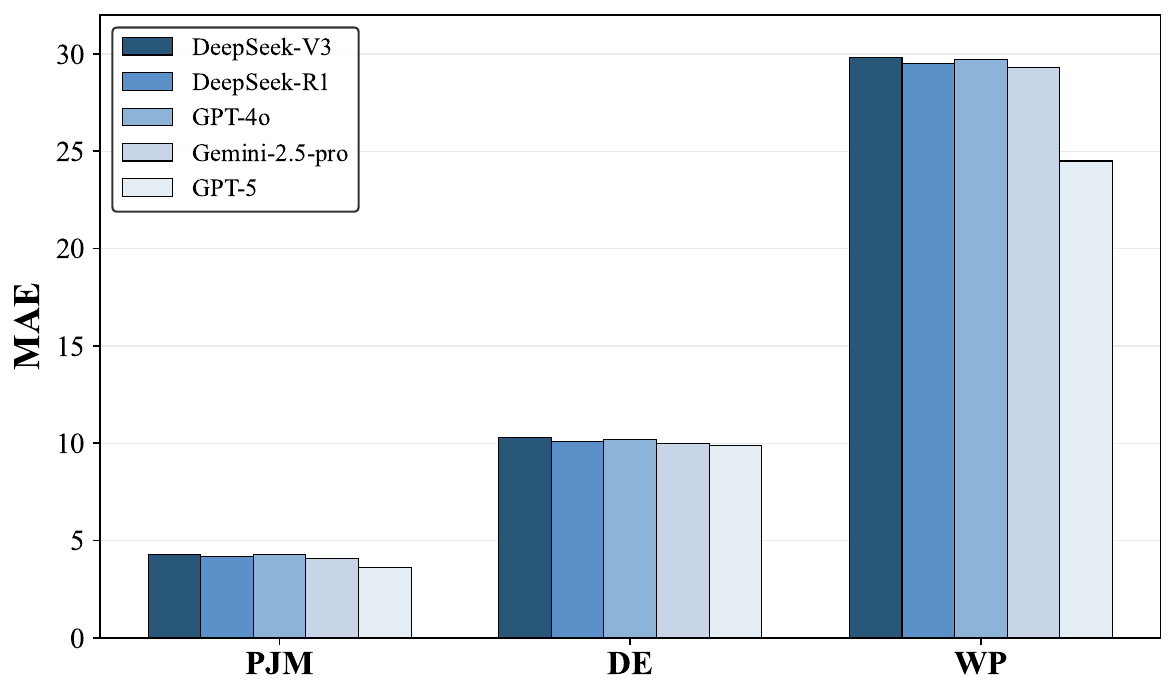}
    \caption{Comparison of forecasting performance across different LLM backbones on multiple datasets. }
    \label{fig:llm_backbone}
\end{figure}

\subsubsection{Ablation Study on Contextual Input.}
To examine how different contextual input components contribute to the contextual input $I_{\text{in}}$, we conduct a component-wise ablation study by removing the data profile, exogenous variables, and timestamp information, respectively. 
As shown in Figure~\ref{fig:ablation_context}, the complete AlphaCast framework consistently yields the lowest MAE, confirming the synergistic contribution of all elements. Specifically, removing timestamps causes significant degradation on periodic datasets such as PJM and SP, highlighting the critical role of explicit temporal identifiers in capturing seasonality that numerical sequences alone cannot convey. Similarly, the exclusion of exogenous variables and static data profile leads to substantial error increases on tasks like ETTm and NP, demonstrating that external drivers and domain-specific inductive biases are essential for reasoning about complex system behaviors beyond historical autoregressive patterns.
\subsubsection{Impact Analysis of the Toolkit}

We remove each toolkit component in turn and report the results in Table~\ref{tab:toolset}. The full model consistently performs best, showing that the four components provide complementary information.
Removing the feature set particularly degrades performance on WP and ETTm, demonstrating the value of statistical and temporal summaries for context preparation. Removing the contextual repository consistently harms performance, confirming the utility of scenario-specific calendar, event, and condition cues. The case library is especially beneficial on DE and ETTm by providing references from similar instances, while the knowledge base substantially improves results on WP, ETTm, and MOPEX through domain-guided adjustment and validation. Further case-library analysis is provided in Appendix~\ref{sec:appendix_case_library}.
Overall, the toolkit combines complementary numerical, contextual, case-based, and domain knowledge to support reasoning-based forecasting.
\begin{figure*}[t!]
    \centering
    \begin{subfigure}[t]{0.48\linewidth}
        \centering
        \includegraphics[width=\linewidth]{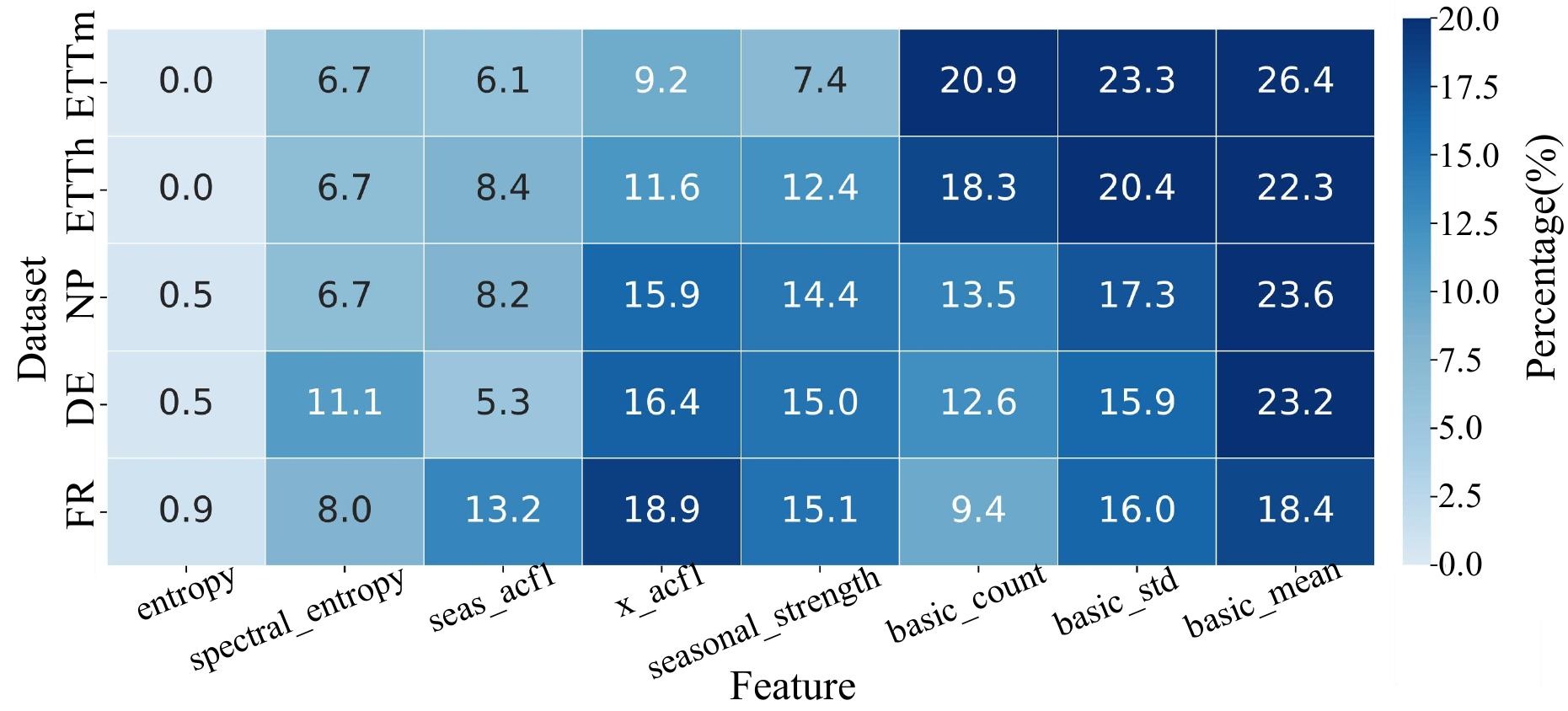}
        \label{fig:feature_case}
    \end{subfigure}
    \hfill
    \begin{subfigure}[t]{0.48\linewidth}
        \centering
        \includegraphics[width=\linewidth]{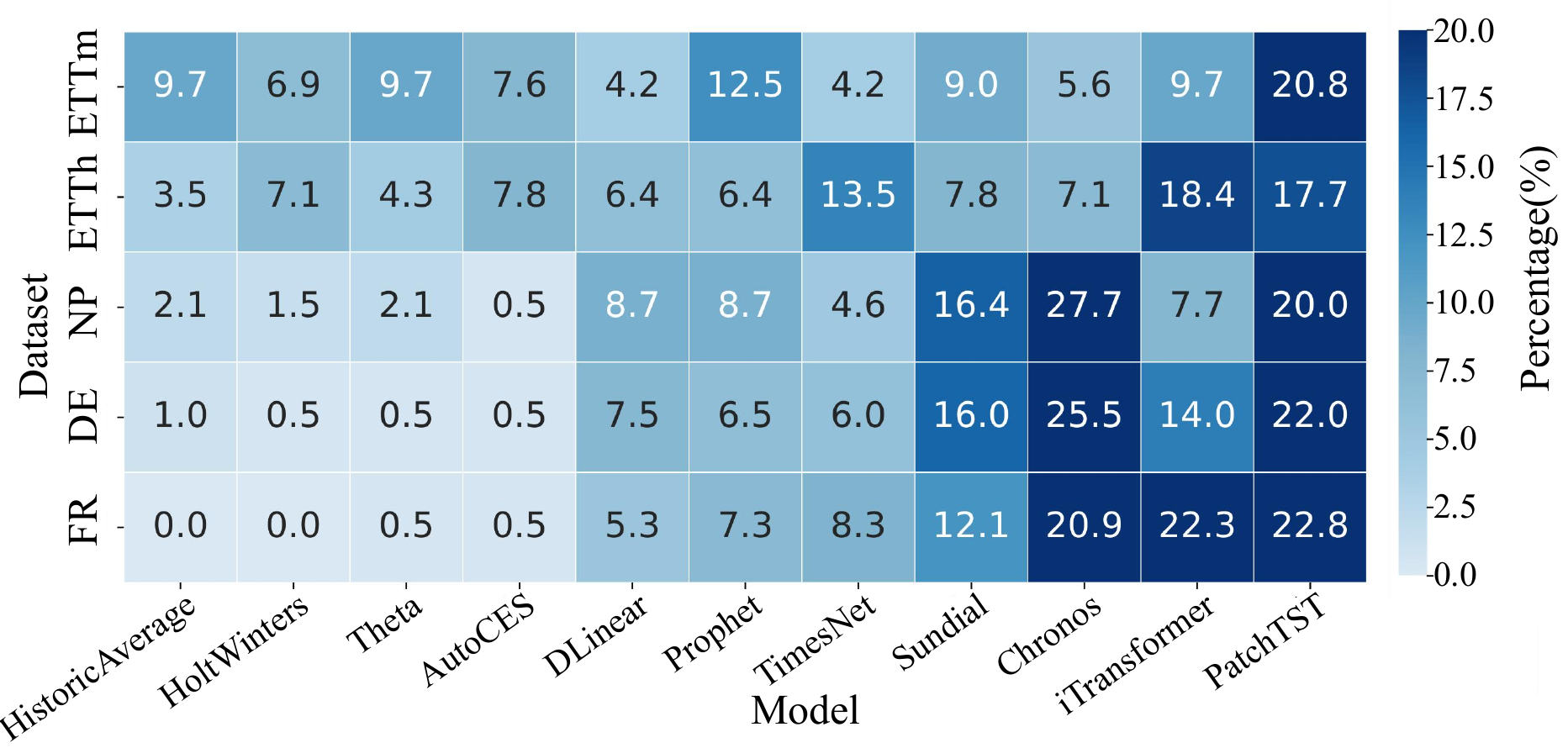}
        \label{fig:model_case}
    \end{subfigure}
    \caption{Feature-level and model-level analysis of AlphaCast across datasets. Selection frequencies (\%). Features: top 8 of 20, normalized within those shown. Models: displayed candidate configurations only, not the full set of candidates.}
    \label{fig:case_study}
\end{figure*}

\input{table/toolset}

\subsection{Exploration Analysis}
We further examine AlphaCast from four perspectives: backbone selection, case-library granularity, generation strategy, and component selection patterns.
\begin{figure*}
    \centering
    \includegraphics[width=\linewidth]{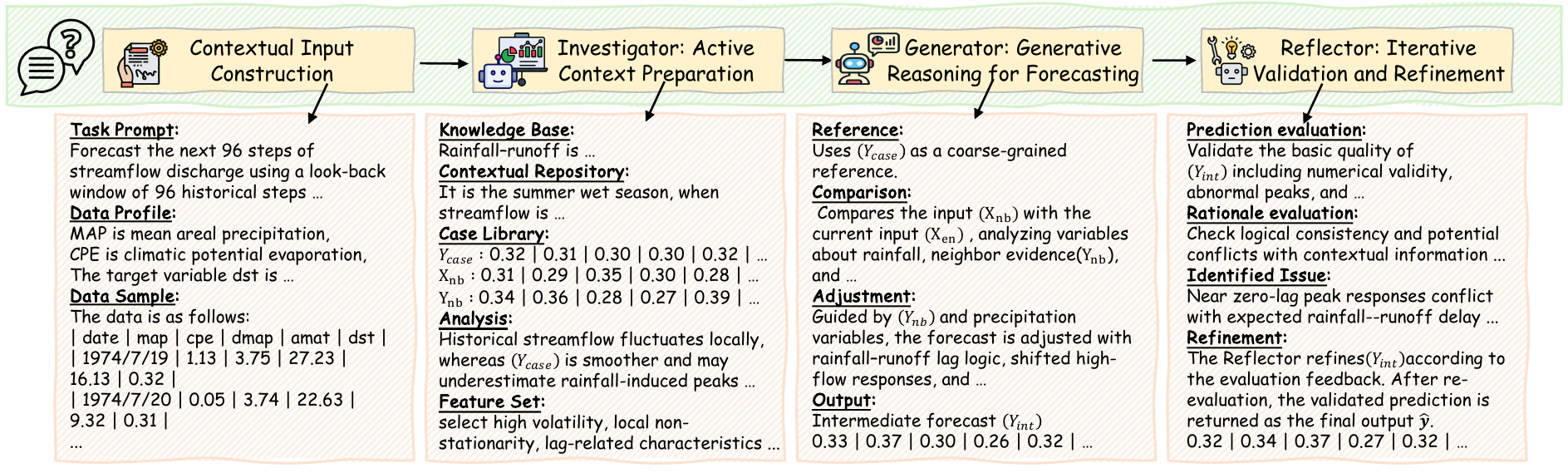}

\caption{A case study of AlphaCast on MOPEX streamflow forecasting, illustrating how contextual preparation, generative reasoning, and reflective validation improve the accuracy and plausibility of the target prediction.}
    \label{fig:case_visualization}
\end{figure*}
\subsubsection{Impact of Forecasting Backbone.}
To investigate the impact of LLM backbones on forecasting performance, we evaluate AlphaCast with different foundation models, including DeepSeek-V3~\cite{liu2024deepseek}, DeepSeek-R1~\cite{guo2025deepseek}, GPT-4o~\cite{achiam2023gpt}, Gemini-2.5-pro~\cite{comanici2025gemini}, and GPT-5~\cite{singh2025openai}. As shown in Figure~\ref{fig:llm_backbone}, GPT-5 achieves the lowest MAE across the evaluated datasets, indicating that the choice of LLM backbone affects the effectiveness of reasoning-based forecasting. The improvement is particularly pronounced on WP, where the prediction window is affected by multiple interacting factors and thus benefits from the rich context organized by AlphaCast. This suggests that stronger LLM backbones can better exploit the structured context provided by AlphaCast, especially when forecasting requires integrating numerical patterns with auxiliary contextual information. Nevertheless, the framework maintains competitive performance across different backbones, showing that the proposed workflow is not tied to a single LLM.

\subsubsection{Impact of Cluster Number.}
\begin{figure}
    \centering
    \includegraphics[width=\linewidth]{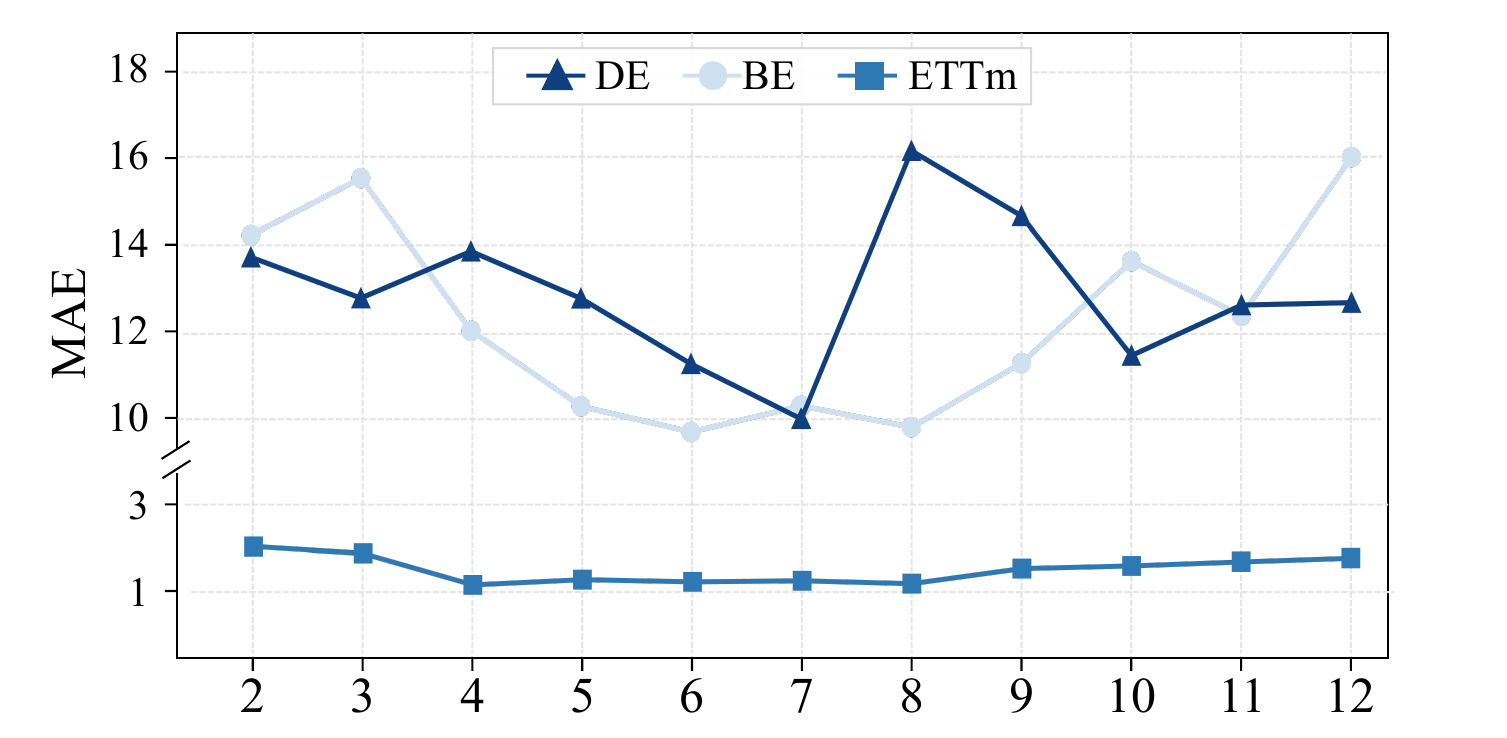}
\caption{Impact of cluster number. Moderate granularity balances specific matching with stable case-based prediction.}
    \label{fig:sensitivity_k}
\end{figure}

To investigate the impact of historical scenario organization, we evaluate forecasting performance across varying cluster counts. As illustrated in Figure~\ref{fig:sensitivity_k}, the error exhibits a U-shaped pattern, initially improving then degrading as the number of clusters increases. Specifically, with a small number of clusters, the library suffers from over-generalization, indiscriminately grouping heterogeneous look-back windows together. Consequently, the synthesized prediction, derived from diverse expert models, becomes a diluted generic signal lacking specificity. Conversely, increasing the cluster count beyond the optimal range leads to excessive fragmentation, resulting in sparse clusters. This sparsity undermines consensus stability, making guidance sensitive to noise. Thus, moderate granularity is crucial for ensuring the accuracy of the prediction.

\subsubsection{Impact of Generation Strategy.}

We compare step-wise generation, rollout generation, and the default whole-horizon generation strategy. Step-wise generation divides the prediction horizon into consecutive segments, whereas rollout generation progressively reuses earlier forecasts as context for later steps. As shown in Table~\ref{tab:generation_strategy}, whole-horizon generation consistently performs best across all datasets. Step-wise generation disrupts temporal dependency continuity, while rollout generation accumulates early prediction errors, particularly on long-horizon or volatile datasets such as WP and BE. By modeling the full prediction window jointly, whole-horizon generation better captures long-term trends, local fluctuations, and seasonal patterns, demonstrating the value of coherent one-shot forecasting.

\subsubsection{Selection Pattern Analysis.}
We analyze the usage frequency within the feature set and case library to understand AlphaCast's selection preferences. As shown in Figure~\ref{fig:case_study}, the heatmaps show that AlphaCast selects different features and candidate models across datasets. In the feature domain, AlphaCast favors fundamental statistical features, such as basic mean, standard deviation, and basic count, suggesting that reliable global properties provide stable evidence for forecasting decisions.
For model selection, AlphaCast shows a clear preference for deep learning architectures, with Transformer-based models and PatchTST frequently selected across datasets, indicating that high-capacity neural backbones provide useful case-based references for capturing temporal dependencies and complex dynamics. Overall, the varying distributions across datasets indicate that AlphaCast adjusts its retrieval strategy according to the input characteristics, rather than applying the same selection pattern across all datasets.

\subsection{Case Study}

We analyze a MOPEX streamflow forecasting case during the summer wet season to illustrate AlphaCast's contextual preparation, generative reasoning, and reflective refinement. Recent precipitation and local fluctuations in the historical discharge series indicate rainfall-driven hydrological dynamics. The case-based reference \(Y_{\text{case}}\) captures the coarse trend but underestimates rainfall-induced peaks. The Investigator selects task-relevant features, including volatility, local non-stationarity, and lag characteristics, to characterize the current scenario. Based on these features, forecast-origin-available precipitation information, and the nearest historical case, the Generator adjusts \(Y_{\text{case}}\) by increasing discharge around periods of stronger rainfall and introducing local peak responses. The Reflector then identifies implausible near-zero-lag peaks and, guided by rainfall--runoff delay priors from the knowledge base, shifts them to better align with hydrological lag behavior. This case illustrates how AlphaCast integrates multi-source evidence, refines coarse references, and enforces domain consistency through reflection.

\input{table/two_stage_comparison}

%% file: table/datasets.tex
\begin{table}[t]
  \centering
  \footnotesize
\caption{Overview of real-world datasets with rich contextual features. These benchmarks span multiple domains and temporal frequencies to ensure comprehensive evaluation.}
  \label{tab:dataset}
  \resizebox{\linewidth}{!}{%
  \begin{tabular}{c c c c c}
    \toprule
    \textbf{Dataset} & \textbf{Domain} & \textbf{Length} & \textbf{Variables} & \textbf{Frequency} \\
    \midrule
    NP            & Electricity Market        & 14,496 & 3  & 1 hour \\
    PJM           & Electricity Market        & 14,496 & 3  & 1 hour \\
    BE            & Electricity Market        & 14,496 & 3  & 1 hour \\
    FR            & Electricity Market        & 14,496 & 3  & 1 hour \\
    DE            & Electricity Market        & 14,496 & 3  & 1 hour \\
    ETTh          & Power Transformer   & 12,240 & 7  & 1 hour \\
    ETTm          & Power Transformer   & 24,096 & 7  & 15 mins \\
    WP   & Wind Power        & 24,096 & 7 & 15 mins \\
    SP   & Solar Power        & 24,096 & 7 & 15 mins \\
    MOPEX         & Hydrology   & 12,240  & 6  & 1 day \\
    \bottomrule
  \end{tabular}
  }
\end{table}

%% file: table/ablation.tex
\begin{table*}[t]
    \centering
    \caption{Ablation study of AlphaCast. ``w/o Toolkit'', ``w/o Reasoning'', and ``w/o Reflection'' remove the forecasting toolkit, reasoning-based generation, and reflective validation, respectively.}
    \label{tab:ablation_full}
    \resizebox{\textwidth}{!}{
    \begin{tabular}{l|cc|cc|cc|cc|cc|cc|cc|cc}
        \toprule
        \multicolumn{1}{c|}{\multirow{2}{*}{Model}}
        & \multicolumn{2}{c|}{PJM}
        & \multicolumn{2}{c|}{BE}
        & \multicolumn{2}{c|}{FR}
        & \multicolumn{2}{c|}{DE}
        & \multicolumn{2}{c|}{SP}
        & \multicolumn{2}{c|}{ETTh}
        & \multicolumn{2}{c|}{ETTm}
        & \multicolumn{2}{c}{MOPEX} \\
        \cmidrule(lr){2-3} \cmidrule(lr){4-5} \cmidrule(lr){6-7} \cmidrule(lr){8-9}
        \cmidrule(lr){10-11} \cmidrule(lr){12-13} \cmidrule(lr){14-15} \cmidrule(lr){16-17}
        & MSE & MAE & MSE & MAE & MSE & MAE & MSE & MAE
        & MSE & MAE & MSE & MAE & MSE & MAE & MSE & MAE \\
        \midrule
        w/o Toolkit
        & 30.76 & 4.01 & 685.41 & 13.06 & 812.56 & 9.94 & 277.12 & 10.83
        & 55.32 & 4.22 & 14.88 & 3.11 & 5.21 & 1.84 & 9.12 & 1.74 \\

        w/o Reasoning
        & 28.99 & 3.97 & 606.53 & 11.24 & 797.26 & 9.13 & 219.87 & 9.98
        & 17.96 & 2.67 & 7.94 & 2.08 & 2.44 & 1.07 & 5.42 & 1.39 \\

        w/o Reflection
        & 25.87 & 3.71 & 551.30 & 11.54 & 741.35 & 8.79 & 235.56 & 11.71
        & 40.18 & 3.39 & 10.95 & 2.84 & 4.01 & 1.58 & 6.02 & 1.65 \\

        \rowcolor{gray!15}
        Full Model
        & \textbf{25.15} & \textbf{3.60}
        & \textbf{536.45} & \textbf{9.61}
        & \textbf{721.02} & \textbf{7.13}
        & \textbf{193.83} & \textbf{9.93}
        & \textbf{13.29} & \textbf{1.84}
        & \textbf{7.64} & \textbf{2.02}
        & \textbf{2.41} & \textbf{1.06}
        & \textbf{4.77} & \textbf{1.35} \\
        \bottomrule
    \end{tabular}
    }
\end{table*}


%% file: table/toolset.tex
\begin{table}[t] 
  \centering
  \setlength{\tabcolsep}{6pt} 
  \caption{Impact analysis of forecasting toolkit components on overall performance, using MSE as the evaluation metric.}
  \resizebox{\columnwidth}{!}{
  \begin{tabular}{c|ccccc}
    \toprule
    \multicolumn{1}{c|}{Datasets} & BE & DE & WP & ETTm & MOPEX \\
    \midrule
    
    w/o Feature Set & 624.27 & 221.97 & 2442.10 & 3.60 & 5.69 \\
    w/o Contextual Repository & 613.26 & 231.13 & 2001.32 & 3.65 & 5.82 \\
    w/o Case Library    & 607.57 & 253.10 & 1670.63 & 3.90 & 4.93 \\
    w/o Knowledge Base  & 641.52 & 211.77 & 2579.32 & 4.49 & 6.57 \\
    \rowcolor{gray!15}
    Full Model & \textbf{536.45} & \textbf{193.83} & \textbf{1548.83} & \textbf{2.41} & \textbf{4.77} \\
    \bottomrule
  \end{tabular}
  }
  \label{tab:toolset}
  
\end{table}

%% file: table/two_stage_comparison.tex
\begin{table}[t]
  \centering
    \caption{Impact of generation strategies. Whole-horizon generation consistently outperforms step-wise and rollout generation across all evaluated datasets.}
  \label{tab:generation_strategy}
  \resizebox{\columnwidth}{!}{
  \begin{tabular}{c| cc cc cc}
    \toprule
    \multicolumn{1}{c|}{Datasets} & \multicolumn{2}{c}{BE} & \multicolumn{2}{c}{PJM} & \multicolumn{2}{c}{WP} \\
    Metric & MSE & MAE & MSE & MAE & MSE & MAE \\
    \midrule
    Step-wise Generation & 616.03 & 10.88 & 31.87 & 4.10 & 2164.37 & 29.83 \\
    Rollout Generation & 597.89 & 11.03 & 30.04 & 3.92 & 2305.87 & 30.18 \\
    Whole-horizon Generation (Ours) & \textbf{536.45} & \textbf{9.61} & \textbf{25.15} & \textbf{3.60} & \textbf{1548.83} & \textbf{24.54} \\
    \bottomrule
  \end{tabular}
  }
\end{table}

%% file: 5-Conclusion.tex
\section{Conclusion}

In this work, we proposed AlphaCast, a human wisdom--LLM intelligence co-reasoning framework that organizes time series forecasting as an expert-like interactive reasoning process with context preparation, generative forecasting, and reflective validation. AlphaCast operationalizes human wisdom through structured forecasting support, including time-series features, contextual cues, case-based references, and domain knowledge, enabling LLM agents to generate and refine predictions with richer contextual grounding. Experiments on short-term and long-term benchmarks show that AlphaCast achieves strong performance compared with diverse forecasting baselines and provides forecast explanations, while ablation analyses validate the contributions of its workflow and toolkit. These results suggest that interactive and context-aware reasoning is a promising direction for time series forecasting.

\section*{Acknowledgments}
This work was supported by grants from the National Natural Science Foundation of China (62502486), the Fundamental Research Funds for the Central Universities of China, USTC Kunpeng-Ascend Scientific and Education Innovation Excellence Center.

%% file: 6-Appendix_new.tex
\section{Case-Based Reference Analysis}
\label{sec:appendix_case_library}
\begin{figure*}[!b]
    \centering
    \includegraphics[width=\linewidth]{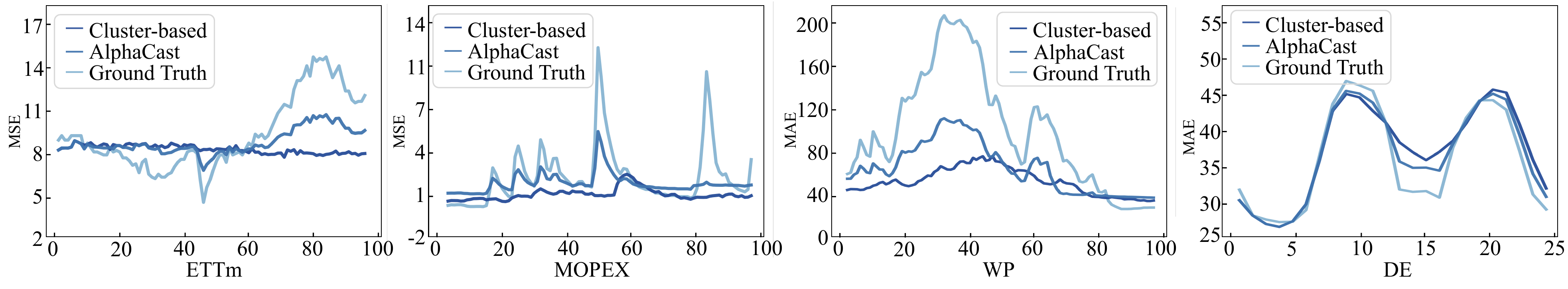}
    \caption{Qualitative comparison of predictions on four datasets (ETTm, MOPEX, WP, and DE). AlphaCast produces predictions closer to the ground truth than the cluster-based prediction, especially under periodic or volatile patterns.}
    \label{fig:case_visualization_appendix}
\end{figure*}

AlphaCast builds a case library consisting of statistical models, deep learning models, and foundation models to provide preliminary forecasting references for its reasoning process. Table~\ref{tab:complete_results} contrasts the standalone performance of each candidate model in the library with the final output of AlphaCast. The results show that AlphaCast mitigates the limitations of individual models through contextual reasoning and iterative refinement, yielding better forecasting performance than individual candidate models on most evaluated settings. Figure~\ref{fig:case_visualization_appendix} shows that the cluster-based reference captures coarse trends but can over-smooth local variations, whereas AlphaCast more closely follows turning points through contextual reasoning and reflective refinement. This complementary role makes the case reference a useful starting point rather than a final forecast. The refinement stage is particularly helpful when local changes are not fully represented by the retrieved historical pattern. Such corrections are especially useful for abrupt changes and context-dependent deviations. They allow AlphaCast to retain the broad historical prior while adapting its forecast to the current prediction scenario, without discarding relevant historical structure.

\input{table/Appendix/complete_result}

\section{Feature Details}
\label{app:feature_set}


We extract a 20-dimensional feature set to summarize each time series to support  reasoning and reflection.

\textit{Statistical features} include seven descriptive statistics: the number of valid observations, mean, standard deviation, minimum, maximum, skewness, and kurtosis. These features characterize basic distributional properties, including sample size, central tendency, dispersion, value range, asymmetry, and tail behavior. They help assess data reliability, scale and variability differences, possible extreme values, and deviations from a symmetric distribution.

\textit{Temporal features}
capture dynamic patterns related to temporal dependence, spectral complexity, local variation, and seasonality. Specifically, we use autocorrelation features before and after differencing to describe persistence and residual temporal dependence, and include seasonal-lag autocorrelation when seasonal information is available. We further use spectral entropy and an alternative entropy estimate to measure the concentration or randomness of the frequency spectrum. In addition, lumpiness, flat spots, and crossing points are used to characterize local variance changes, flat or discretized segments, and oscillatory behavior. A seasonal strength proxy is finally included to summarize the degree of seasonality.

%% file: table/Appendix/complete_result.tex
\begin{table*}[!b]
\centering
\small
\caption{Forecasting performance on short-term and long-term benchmarks. We compare AlphaCast with each individual candidate model in the case library to show the benefit of interaction-driven reasoning and iterative refinement.}
\label{tab:complete_results}
\setlength{\tabcolsep}{2.5pt}
\renewcommand{\arraystretch}{1.15}
\resizebox{\textwidth}{!}{
\begin{tabular}{c|cc cc cc cc cc|cc cc cc cc cc}
\toprule
\cmidrule(r){1-1} \cmidrule(lr){2-11} \cmidrule(l){12-21}
\multicolumn{1}{c|}{Dataset}
& \multicolumn{2}{c}{NP}
& \multicolumn{2}{c}{PJM}
& \multicolumn{2}{c}{BE}
& \multicolumn{2}{c}{FR}
& \multicolumn{2}{c|}{DE}
& \multicolumn{2}{c}{ETTh}
& \multicolumn{2}{c}{ETTm}
& \multicolumn{2}{c}{WP}
& \multicolumn{2}{c}{SP}
& \multicolumn{2}{c}{MOPEX} \\
\multicolumn{1}{c|}{Metric}
& MSE & MAE
& MSE & MAE
& MSE & MAE
& MSE & MAE
& MSE & MAE
& MSE & MAE
& MSE & MAE
& MSE & MAE
& MSE & MAE
& MSE & MAE \\
\midrule

HistoricAverage
& 54.43 & 5.23
& 108.61 & 8.10
& 745.45 & 16.74
& 955.66 & 14.63
& 381.71 & 14.57
& 10.31 & 2.57
& 3.30 & 1.31
& 2671.84 & 36.53
& 59.76 & 6.13
& 6.14 & 1.71 \\

SNaive
& 44.89 & 4.10
& 41.39 & 4.65
& 857.12 & 13.70
& 915.80 & 11.15
& 415.72 & 13.36
& 10.75 & 2.47
& 2.75 & 1.19
& 1948.06 & \underline{27.88}
& 79.61 & 4.56
& 7.18 & 1.62 \\

ARIMA
& 55.51 & 4.45
& 33.02 & 4.00
& 869.44 & 17.04
& 964.00 & 13.22
& 372.91 & 11.82
& 10.88 & 2.43
& 2.71 & 1.17
& 2166.64 & 28.77
& 80.48 & 4.60
& 9.19 & 1.85 \\


Theta
& 60.11 & 4.78
& 36.07 & 4.30
& 750.76 & 13.84
& 855.15 & 11.28
& 424.50 & 12.53
& 11.77 & 2.54
& 3.35 & 1.26
& 2277.53 & 29.88
& 79.98 & 4.56
& 8.06 & 1.78 \\

Holt-Winters
& 52.30 & 4.49
& 64.06 & 5.06
& 874.59 & 14.73
& 1021.85 & 12.84
& 366.36 & 11.44
& 71.12 & 4.58
& 16.74 & 2.57
& 41607.96 & 107.58
& 131.50 & 5.49
& 1464.18 & 7.17 \\


CES
& 40.78 & 3.99
& 38.33 & 4.47
& 808.61 & 14.41
& 1084.71 & 12.14
& 315.58 & 11.45
& 24.22 & 3.65
& 4.12 & 1.48
& 9364.98 & 42.04
& 78.71 & 4.55
& 6.40 & 1.47 \\

Prophet
& 57.79 & 5.00
& 52.26 & 5.53
& 992.90 & 16.94
& 1035.70 & 14.42
& 320.63 & 13.25
& 46.70 & 4.59
& 21.18 & 3.11
& 8071.24 & 56.06
& 64.81 & 6.44
& 12.39 & 2.50 \\

Autoformer
& 46.10 & 4.86
& 77.57 & 6.43
& 890.84 & 16.62
& 934.36 & 13.36
& 331.31 & 13.33
& 11.60 & 2.64
& 4.22 & 1.57
& 2567.61 & 37.00
& 39.13 & 4.34
& 6.17 & 1.96 \\

DLinear
& 32.22 & 3.84
& 42.15 & 4.64
& 658.53 & 12.83
& 811.45 & 10.45
& 239.93 & 12.83
& 8.51 & 2.24
& 2.63 & 1.14
& 1932.20 & 29.32
& 19.06 & 2.73
& 5.26 & \underline{1.33} \\

TimesNet
& 32.12 & 3.81
& 34.89 & 4.34
& 636.66 & 11.30
& 929.10 & 11.10
& 209.37 & 10.20
& \underline{7.94} & \underline{2.08}
& \underline{2.44} & \underline{1.07}
& \underline{1930.88} & 29.40
& 18.43 & 2.78
& 4.96 & 1.35 \\

PatchTST
& \underline{24.63} & 3.27
& 31.87 & 4.17
& 627.15 & 11.44
& \underline{797.26} & 9.13
& \underline{208.89} & 10.04
& 8.40 & 2.17
& 2.62 & 1.11
& 2269.64 & 32.46
& 19.43 & 2.78
& 5.36 & 1.44 \\

iTransformer
& 27.09 & 3.43
& 35.13 & 4.37
& \underline{606.53} & 11.24
& 940.23 & 11.09
& 229.96 & 10.23
& 8.31 & 2.14
& 3.14 & 1.28
& 2063.99 & 29.89
& \underline{17.96} & \underline{2.67}
& \underline{4.88} & 1.34 \\

Chronos
& \textbf{22.18} & \underline{3.18}
& \underline{25.70} & \underline{3.63}
& 625.63 & \underline{9.62}
& 803.08 & \underline{7.54}
& 223.15 & \underline{10.00}
& 9.40 & 2.25
& 2.75 & 1.18
& 2268.52 & 29.62
& 79.25 & 4.57
& 5.28 & \underline{1.31} \\

Sundial
& 28.67 & 3.36
& 30.70 & 3.99
& 651.24 & 11.00
& 942.08 & 8.54
& 264.62 & 10.71
& 9.44 & 2.31
& 3.21 & 1.26
& 2167.94 & 31.22
& 100.30 & 6.69
& 5.15 & 1.42 \\

\midrule
\rowcolor[gray]{0.95}
\textbf{AlphaCast}
& 27.11 & \textbf{3.15}
& \textbf{25.15} & \textbf{3.60}
& \textbf{536.45} & \textbf{9.61}
& \textbf{721.02} & \textbf{7.13}
& \textbf{193.83} & \textbf{9.93}
& \textbf{7.64} & \textbf{2.02}
& \textbf{2.41} & \textbf{1.06}
& \textbf{1548.83} & \textbf{24.54}
& \textbf{13.29} & \textbf{1.84}
& \textbf{4.77} & 1.35 \\

\bottomrule
\end{tabular}
}
\end{table*}